\documentclass{article}
\usepackage{spconf,amsmath,epsfig}
\usepackage{graphicx}
\usepackage{amssymb, url}
\usepackage{multirow, tikz, makecell, verbatim, booktabs, array}
\usepackage[labelsep=period]{caption}
\usepackage[labelformat=simple]{subcaption}

\DeclareMathOperator*{\argmax}{argmax}

\let\OLDthebibliography\thebibliography
\renewcommand\thebibliography[1]{
  \OLDthebibliography{#1}
  \setlength{\parskip}{0pt}
  \setlength{\itemsep}{0pt plus 0.3ex}
}

\begin{document}\sloppy
\topmargin=0mm
\def\x{{\mathbf x}}
\def\L{{\cal L}}

\title{BALANCING DOMAIN EXPERTS FOR LONG-TAILED CAMERA-TRAP RECOGNITION}
%
\name{Byeongjun Park, Jeongsoo Kim, Seungju Cho, Heeseon Kim, Changick Kim}
\address{\small{School of Electrical Engineering, KAIST, Daejeon, Republic of Korea}\\
\small{\{pbj3810, jngsoo711, joyga, hskim98, changick\}@kaist.ac.kr}}

\maketitle

\newcommand{\etal}{\emph{et al}.\@ }

\begin{abstract}
Label distributions in camera-trap images are highly imbalanced and long-tailed, resulting in neural networks tending to be biased towards head-classes that appear frequently.
Although long-tail learning has been extremely explored to address data imbalances, few studies have been conducted to consider camera-trap characteristics, such as multi-domain and multi-frame setup.
Here, we propose a unified framework and introduce two datasets for long-tailed camera-trap recognition.
We first design domain experts, where each expert learns to balance imperfect decision boundaries caused by data imbalances and complement each other to generate domain-balanced decision boundaries.
Also, we propose a flow consistency loss to focus on moving objects, expecting class activation maps of multi-frame matches the flow with optical flow maps for input images.
Moreover, two long-tailed camera-trap datasets, WCS-LT and DMZ-LT, are introduced to validate our methods.
Experimental results show the effectiveness of our framework, and proposed methods outperform previous methods on recessive domain samples.
\end{abstract}

\section{Introduction}
\label{sec:intro}
Biologists and ethologists often use camera-traps to capture animals inconspicuously to study the population biology and dynamics \cite{burton2015wildlife}.
While these cameras automatically collect massive data, identifying species by humans is time-consuming and labor-intensive, limiting research productivity.
Therefore, deep neural networks \cite{simonyan2014very, he2016deep} have recently received attention for their ability to automate the identification process, making camera-trap studies scalable \cite{go2021fine, norouzzadeh2018automatically}.
Nevertheless, neural networks tend to be biased towards the species that frequently appear, limiting studies that require diverse animal species, specifically on endangered species.  

Early camera-trap recognition methods focus on long-tailed recognition to make the neural network more tail-sensitive, and prevailing methods are summarized as follows: Re-weighting the loss \cite{cui2019class, cao2019learning}; Re-sampling the data for minor classes \cite {zou2018unsupervised}; Transfer learning \cite{liu2019large}. Recently, multi-expert networks have achieved considerable successes in long-tailed recognition by forcing experts to learn each classifier for different sub-groups in parallel \cite{wang2021longtailed, cai2021ace}.

\begin{figure}[!t]
\centering
\includegraphics[width=\linewidth]{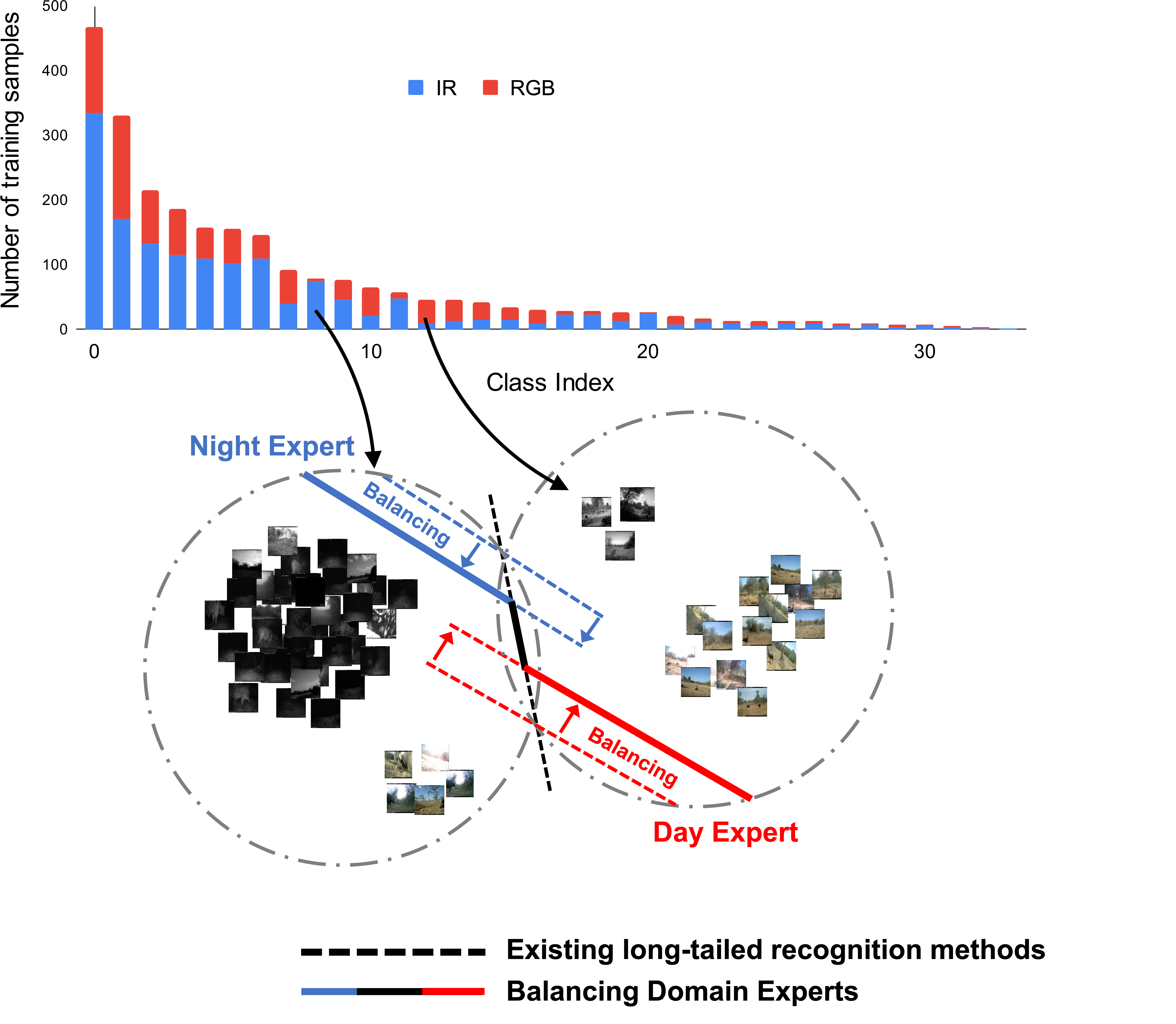}
\vspace{-0.6cm}
\caption{Two classes are highly imbalanced between two domains. Existing long-tailed recognition methods resolve the imbalanced label distribution (black dotted line). However, the discriminability of recessive domain features is not sufficiently improved due to data imbalances, resulting in samples being located near the boundary. Consequently, IR boundaries (blue dotted line) and RGB boundaries (RGB dotted line) are still biased towards the head-class of each domain. Thus, we propose that domain experts balanced by a simple re-weighting method give a margin for tail-classes of each domain, and experts complement each other to generate domain-balanced decision boundaries (blue-black-red solid line).}
\vspace{-0.4cm}
\label{fig:figure1}
\end{figure}

\begin{figure*}[!t]
\centering
\includegraphics[width=0.85\linewidth]{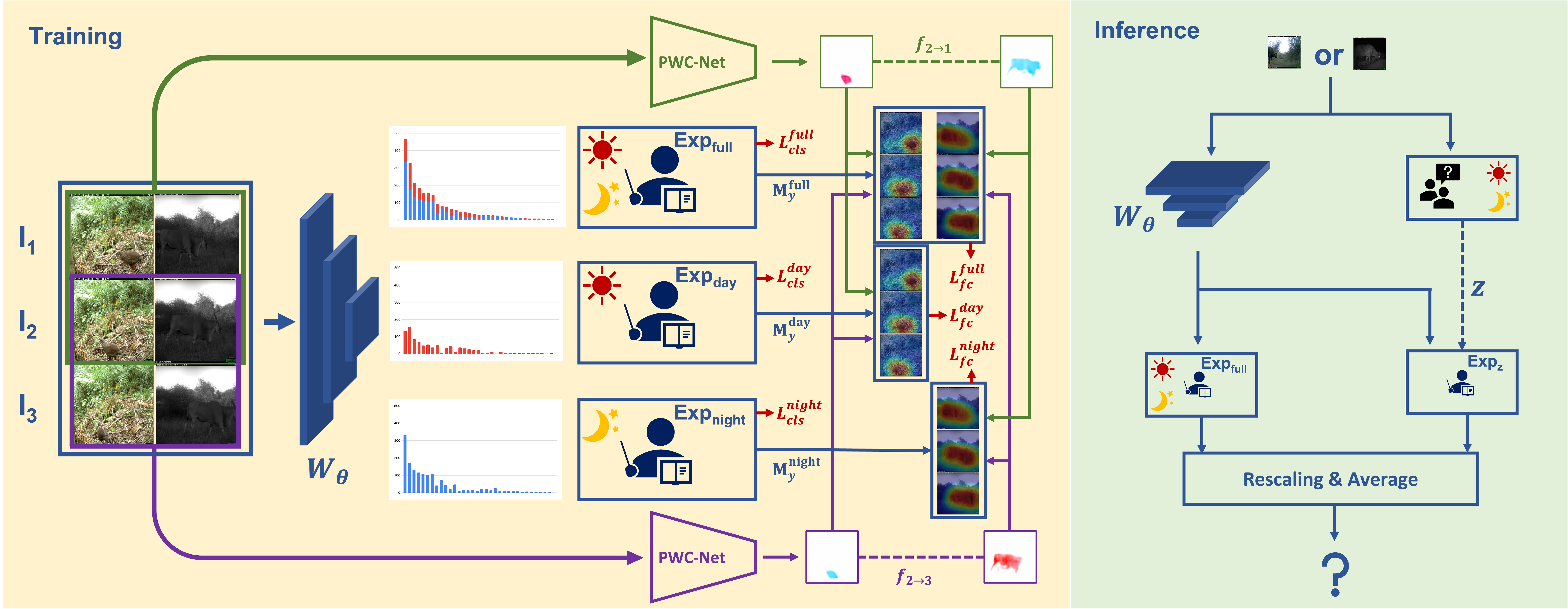}

\caption{Network architecture for training and inference. In the training scheme, batches are processed to estimate the past flow $f_{2 \to 1}$ and the future flow $f_{2 \to 3}$, and also processed to feature extractor $W_{\theta}$. The output feature is then processed into each expert $\mbox{EXP}_{i}$ (i.e., $\psi_{i}$) and generate class activation map $M^{i}_y$. The classification loss $L^{i}_{cls}$ is applied for each expert, and the flow consistency loss $L^{i}_{fc}$ is applied on each expert with $f_{2 \to 1}$ and $f_{2 \to 3}$. In the inference time, a single input image is processed into $W_{\theta}$, and the full-domain expert and the corresponding expert complement each other to output the label.}
\vspace{-0.4cm}
\label{fig:network}
\end{figure*}

Despite these efforts to prefer tail-classes, limited efforts have been made to address the data imbalance between domains when images are acquired from multiple domains with different label distributions.
Especially in camera-trap images, the samples of diurnal (e.g., \emph{marten}) and nocturnal (e.g., \emph{raccoon}) animals are biased in the corresponding domain, respectively, resulting in the previous methods being often biased towards the dominant domain. Therefore, the boundary of the samples in the recessive domain may have the potential to shrink, degrading the classification performance.

In this paper, we propose domain experts that mitigate the bias by combining decision boundaries, where domain experts are separately learned from each domain.
There are two types of experts, the sub-domain expert and the full-domain expert.
Exclusive sub-domain experts, one is for the night (i.e., IR) and the other is for the day (i.e., RGB), are individually specialized in that domain, and the focal loss \cite{lin2017focal} is applied to balance the imperfect decision boundary caused by the data imbalance.
The full-domain expert learns from all input images since IR and RGB images are essential for learning object boundaries and contextual information, respectively.
The full-domain expert and two sub-domain experts complement each other to create better domain-balanced decision boundaries, and details are shown in Fig. \ref{fig:figure1}.

While previous methods treat successive images taken by the camera-trap as independent images, we further propose a flow consistency loss for each expert to leverage the multi-frame information.
We regulate the class activation map of multi-frames following the optical flow map estimated from pre-trained PWC-Net \cite{sun2018pwc}. Thus, the flow consistency loss enhances experts to pay more attention to moving objects. 

To validate our method, we introduce two camera-trap datasets, WCS-LT and DMZ-LT, which are multi-domain and multi-frame with long-tailed distributions. In addition, we evaluate the accuracy on these datasets and show that our method outperforms the previous methods for samples from the recessive domain as well as the dominant domain.

\section{METHOD}
\subsection{Network Architecture}

The architecture of the proposed network is shown in Fig. \ref{fig:network}, and multiple experts are trained in parallel with a shared backbone. 
We use three consecutive frames as an input sequence, and details are described in Section \ref{section: dataset}.
Existing classifiers tend to perform better on the dominant domain samples than on the recessive domain samples; however, domain experts can mitigate the bias.
Therefore, we design domain experts consisting of the full-domain expert and the $K$ sub-domain experts. 
We fix $K=2$ to treat \emph{day} and \emph{night} domain in the camera-trap setup.
These two types of experts are complemented each other from two aspects: (1) The full-domain expert learns valuable information from both domains and makes robust predictions but is biased towards dominant domain samples; (2) Sub-domain experts support the full-domain expert to predict without prejudices and give confidence to the prediction.

\subsection{Training Scheme}

In this section, we briefly illustrate the training scheme for a input sequence $\mathcal{S}=\left\{I_{1}, I_{2}, I_{3}\right\}$ and experts $\Psi=\left\{\psi_{full}, \psi_{day}, \psi_{night}\right\}$. Here, the domain set and the class set are defined as $\mathbb{D}=\left\{ \mbox{day}, \mbox{night} \right\}$ and $\mathbb{C}=\left\{1, 2, \cdots, C \right\}$, respectively. 
The domain $z \in \mathbb{D}$ of $S$ is determined by ensuring that the input values of each channel are identical, as IR images are gray-scale.
We denote $y \in \mathbb{C}$ as the class label of $S$, and determine whether the inputs are majority samples (MJs) or minority samples (MNs) depending on whether $z$ is the dominant domain of $y$.

While $\psi_{full}$ uses all input sequences, each sub-domain expert uses the sequences of the corresponding domain.
With this data split mechanism, $\psi_{day}$ and $\psi_{night}$ learn the domain-specific decision boundaries without being hindered by data imbalances between domains.
Following \cite{cai2021ace, kang2019decoupling}, we use ResNet-50 \cite{he2016deep} as a backbone and define each expert $\psi_{i} \in \Psi$ as a residual block followed by a global average pooling layer and a learnable weight scaling classifier.
Consequently, output logits before SoftMax operation of $\psi_{i}$ are $\textbf{x}_{i, 1}, \textbf{x}_{i, 2}, \textbf{x}_{i, 3} \in \mathbb{R}^{1 \times C }$.
To avoid interfering with each other's learning, loss functions are applied to the experts separately.
First, we use the focal loss \cite{lin2017focal} for $\psi_{i}$ as the classification loss as
\begin{equation}
L^{i}_{cls} = -\sum^{3}_{j=1}{(1-\sigma(\textbf{x}_{i, j})_{y})^{\gamma}\log(\sigma(\textbf{x}_{i, j})_{y})},
\label{eq:cls}
\end{equation}
where $\sigma(\textbf{x}_{i, j})_{y} \in \mathbb{R}$ is the output logit of the class $y$ after the SoftMax operation for the input logit $\textbf{x}_{i, j}$, and we fix $\gamma = 5$.

To further increase the discriminability of each expert, a flow consistency loss is proposed to make flow-consistent experts expect to pay more attention to moving objects.
We apply the flow consistency loss for the class activation map of each expert, where the class activation map of multiple frames to have a flow-consistent with the optical flow map estimated in the pre-trained PWC-Net \cite{sun2018pwc}.

We first extract the class activation map $M^{ij}_{y}$ for the class label $y$ with the $j$-th frame and $\psi_{i}$ as
\begin{equation}
    M^{ij}_{y} = \sum_{k}{w^{i}_{k, y} A^{ij}_{k}},
\end{equation}
where  $w^{i}_{k, y} \in \mathbb{R}$ is the fully-connected layer's weight of $\psi_{i}$ at the $k$-th row and the $y$-th column, and $A^{ij}_{k}$ is the $k$-th channel of the feature map at the last convolution layer of $\psi_{i}$ for the $j$-th frame.
In the context of \cite{kang2019decoupling}, we freeze the feature map to allow the gradient back-propagates only to the fully-connected layer. 

With two flow maps estimated from the pre-trained PWC-Net, a past flow map $f_{2\to 1}$ and a future flow map $f_{2\to 3}$, we generate warped maps $\hat{M}^{i, 1}_{y}$ and $\hat{M}^{i, 3}_{y}$ from $M^{i, 1}_{y}$ and $M^{i, 3}_{y}$, respectively. Then, the flow consistency loss is applied for $\psi_{i}$ to match the warped maps with $M^{i, 2}_{y}$ as
\begin{equation}
L^{i}_{fc} = L_{ph}(M^{i, 2}_{y}, \hat{M}^{i, 1}_{y}) + L_{ph}(M^{i, 2}_{y}, \hat{M}^{i, 3}_{y}),
\end{equation}
where $L_{ph}$ is a photometric consistency loss which is commonly used for self-supervised optical flow and depth estimation tasks \cite{jonschkowski2020matters, godard2019digging} as
\begin{equation}
L_{ph}(a, b) = \frac{\alpha}{2} ({1 - SSIM_{a, b}}) + (1 - \alpha){ ||a - b||}_1.
\end{equation}
Here, we fix $\alpha = 0.85$, and $SSIM_{a, b}$ is the structure similarity \cite{wang2004image} between $a$ and $b$. 
With a weight for the flow consistency loss $\beta=0.02$, the overall loss function for $\psi_{i}$ is defined as
\begin{equation}
L^{i} = L^{i}_{cls} + \beta L^{i}_{fc}.
\label{eq:overall}
\end{equation}

\subsection{Inference Scheme}
Different from the training phase, inferences are made on one image, considering the camera-trap only captures a single image.
Depending on the domain $z$ of the input data, we use a full-domain expert $\psi_{full}$ and a sub-domain expert $\psi_{z}$. Similar to \cite{cai2021ace}, the output logit $\textbf{x}_{z} \in \mathbb{R}^{1 \times C}$ of $\psi_{z}$ is modified to be ${\tilde{\textbf{x}}}_{z}$ by the l2-norm of the fully-connected layer's weights as

\begin{equation}
{\tilde{\textbf{x}}}_{z} = \frac{\sqrt{\sum_{k}{\sum_{c \in \mathbb{C}}{({w^{z}_{k, c}})^2}}}}{\sqrt{\sum_{k}{\sum_{c \in \mathbb{C}}{({w^{full}_{k, c}})^2}}}} \cdot \textbf{x}_{z}.
\end{equation}
Then, the modified output logit is averaged over two experts as 
\begin{equation}
\textbf{x} = \frac{\textbf{x}_{full} + \tilde{\textbf{x}}_{z}}{2},
\end{equation}
and the estimated category is defined as
\begin{equation}
y_{pred} = \argmax_{c \in \mathbb{C} }{(\sigma(\textbf{x})_{c})}.
\end{equation}

\section{Camera-Trap Datasets}
\label{section: dataset}

\begin{figure}[htb!]
\begin{subfigure}{.23\textwidth}
  \centering
  \includegraphics[width=\textwidth]{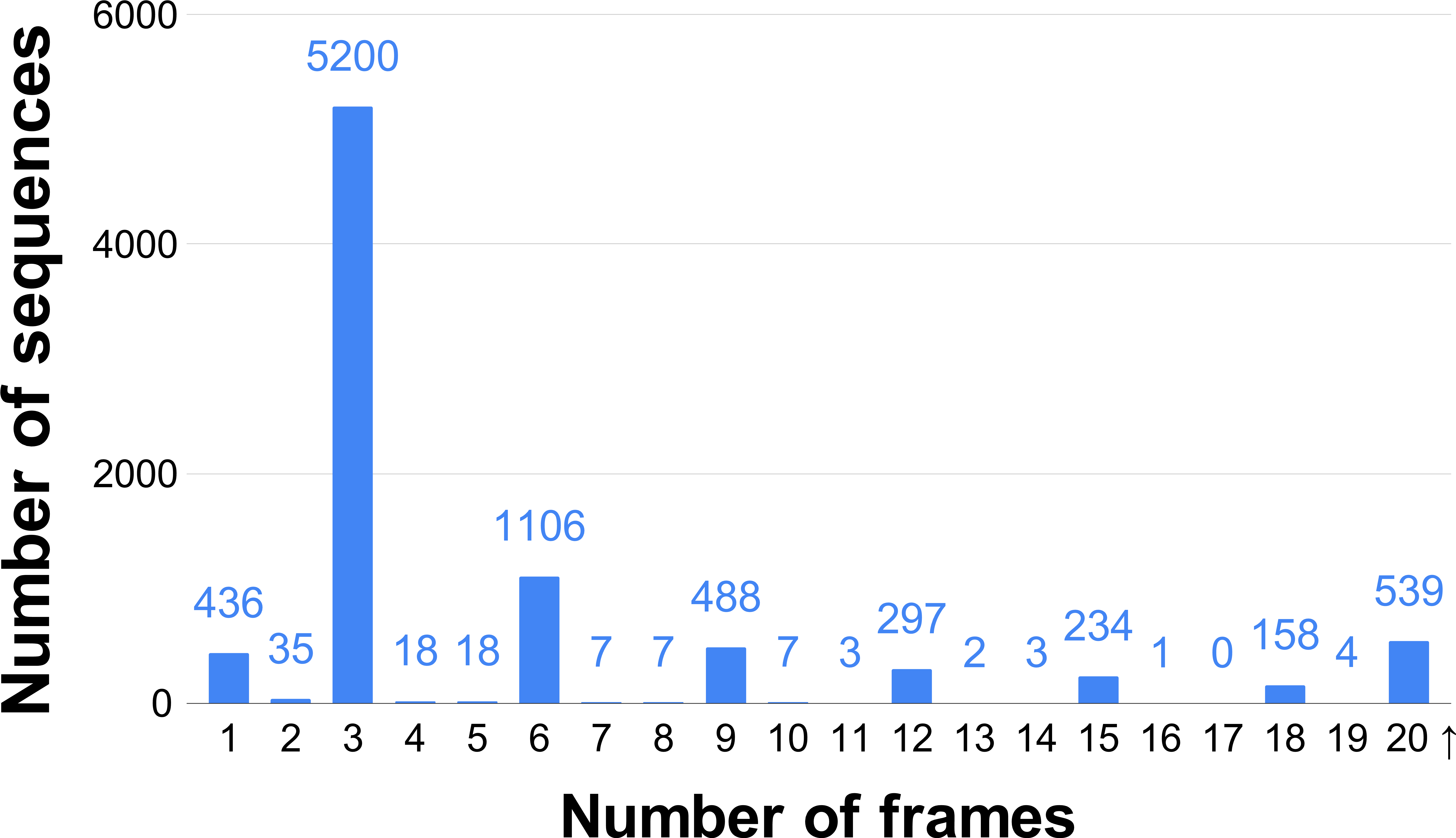}
  \vspace{-0.2cm}
  \caption{}
  \label{fig:WCSa}
\end{subfigure}
\hfill
\begin{subfigure}{.23\textwidth}
  \centering
  \includegraphics[width=\textwidth]{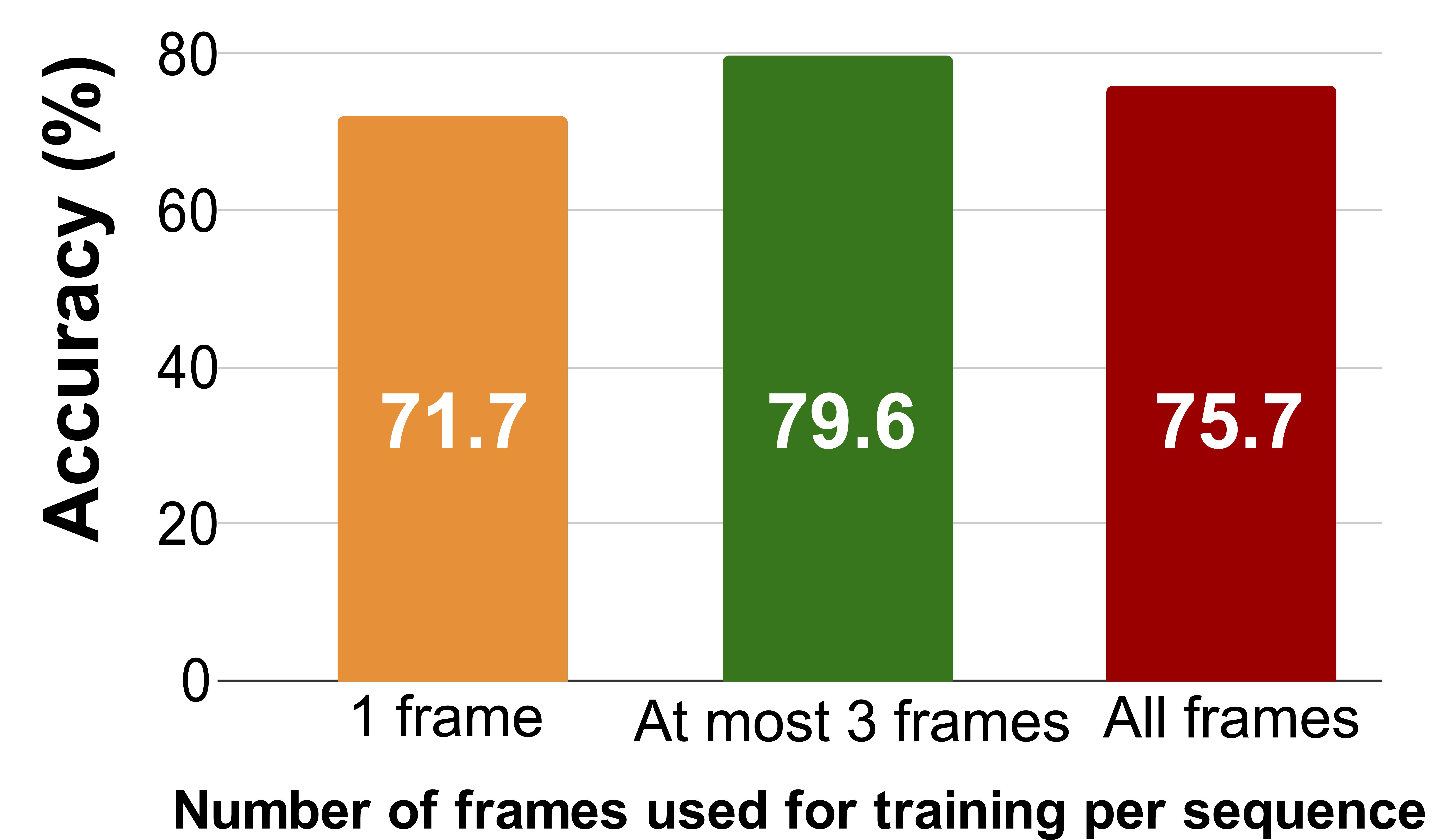}
  \vspace{-0.2cm}
  \caption{}
  \label{fig:WCSb}
\end{subfigure}

\begin{subfigure}{.46\textwidth}
  \centering
  \includegraphics[width=\textwidth]{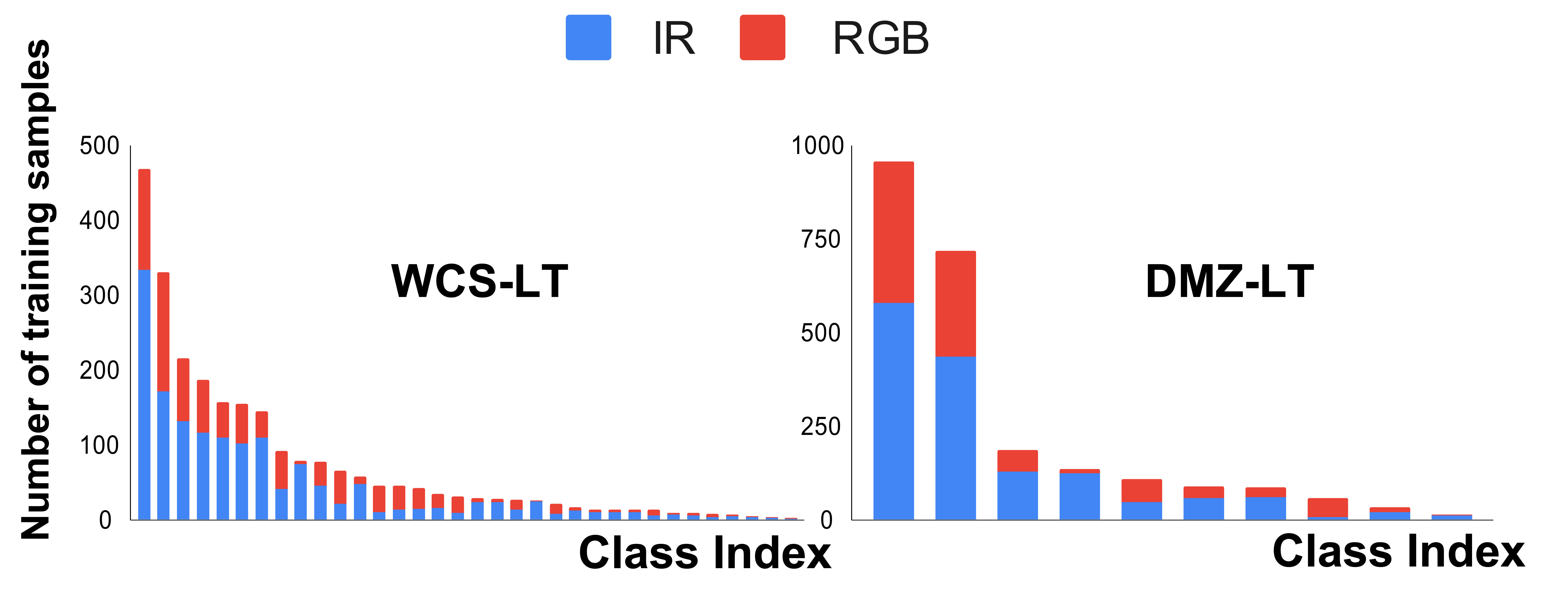}
    \vspace{-0.4cm}
  \caption{}
  \label{fig:WCSc}
\end{subfigure}
\caption{iWildCAM2020 \cite{beery2020iwildcam} statistics of (a) The number of frames per a sequence, (b) Accuracy according to the number of frames used. (c) Long-tailed label distributions of WCS-LT (left) and DMZ-LT (right).}
\vspace{-0.2cm}
\label{fig:WCS}
\end{figure}

\begin{table*}[t]
\centering
\resizebox{1.0\linewidth}{!}{
\begin{tabular}{c||c|ccc|ccc|ccc|ccc}
\Xhline{3\arrayrulewidth}
\multirow{3}{*}{Dataset} & \multirow{3}{*}{Method}  & \multicolumn{10}{c}{Top-1 Accuracy (\%)} \\ \cline{3-12}
                         &                          & \multirow{2}{*}{Many} & \multirow{2}{*}{Medium} & \multirow{2}{*}{Few} & \multicolumn{3}{c|}{Major} & \multicolumn{3}{c|}{Minor}  & \multirow{2}{*}{All}\\ 
                         &                          &                       &                         &                      & Balance & Imbalance & Total       & Balance   & Imbalance & Total       &                        \\ \hline \hline
\multirow{6}{*}{WCS-LT}&baseline (ResNet-50)     & 88.0 & 60.6 & 36.9 & 79.3 & 66.5 & 77.9 & 81.3 & 59.8 & 78.8 & 78.4 \\
&Focal loss \cite{lin2017focal} & 89.7 & 62.2 & 39.9 & \textbf{82.2} & 67.9 & 80.6 & 82.4 & 57.6 & 79.6 & 80.1 \\
&CB loss \cite{cui2019class} & 89.2 & 58.9 & 36.9 & 80.0 & 70.5 & 79.0 & 81.0 & 58.9 & 78.5 & 78.7 \\
&LDAM+DRW \cite{cao2019learning} & 88.9 & 62.1 & 44.4 & 80.1 & 67.9 & 78.7 & 83.1 & 61.2 & 80.7 & 79.7 \\
&ACE (3 experts) \cite{cai2021ace}    & 80.4 & 59.9 & \textbf{62.6} & 74.8 & 69.6 & 74.2 & 77.0 & 52.7 & 74.2 & 74.2 \\
&\textbf{Ours}              & \textbf{89.8} & \textbf{66.6} & 52.0 & \textbf{82.2} & \textbf{75.5} & \textbf{81.4} & \textbf{84.6} & \textbf{64.7} & \textbf{82.4} & \textbf{81.9} \\ \hline \hline
\multirow{6}{*}{DMZ-LT}&baseline (ResNet-50)     & 50.0 & 59.6 & - & 50.6  & 89.9 & 51.7 & 51.8 & 37.7 & 51.4 & 51.5 \\
&Focal loss \cite{lin2017focal} & 48.8 & 59.8 & - & 49.4 & 88.4 & 50.4 & 51.1 & 39.1 & 50.7 & 50.6 \\
&CB loss \cite{cui2019class} & 51.1 & 45.1 & - & 51.0 & 78.3 & 51.7 & 49.7 & 14.5 & 48.8 & 50.2 \\
&LDAM+DRW \cite{cao2019learning} & 52.2 & \textbf{65.1} & - & 57.1 & \textbf{91.3} & 58.0 & 50.6 & 46.4 & 50.5 & 54.2 \\
&ACE (3 experts) \cite{cai2021ace}    & \textbf{64.9} & 42.7 & - & 54.9 & 87.0 & 55.7 & 50.5 & 31.9 & 50.0 & 52.9 \\ 
&\textbf{Ours}  & 56.6 & 62.9 & - & \textbf{61.4} & 81.2 & \textbf{61.9} & \textbf{53.0} & \textbf{65.2} & \textbf{53.3} & \textbf{57.6} \\ \Xhline{3\arrayrulewidth}
\end{tabular}%
}
\caption{Comparison on long-tailed camera-trap datasets, WCS-LT and DMZ-LT. We show that our proposed method outperforms the previous methods in most evaluation metrics. Best results in each metric are in \textbf{bold}. Note that we except the accuracy for the few-shot split in the DMZ-LT dataset since there is one category that belongs to the few-shot split.}
\vspace{-0.3cm}
\label{table:compare}
\end{table*}

\noindent
We explore the relationship between the number of frames in a sequence and the classification accuracy since iWildCAM2020 \cite{beery2020iwildcam} provides the frame information. Given that camera-traps capture images during the object moves, Fig. \ref{fig:WCSa} shows that most sequences consisting of up to three frames capture dynamic objects, and the rest of the sequences capture barely moving objects. Also, we observe that neural networks overfit to these redundant frames. Figure \ref{fig:WCSb} shows the classification performance increases as more images are used rather than one image per sequence, and the best performance is when the first three frames are used for training while the performance deteriorates when all frames are used. 

While existing camera-trap datasets \cite{beery2020iwildcam, van2018inaturalist} consider consecutive frames as independent images, and also disregard prior knowledge for each domain, we introduce two benchmarks to cover general camera-trap settings with three characteristics: (1) Training on multi-frame sequences and testing with a single image; (2) Multi-domain with different long-tailed label distributions; (3) Domain-Balanced test dataset.

\noindent
\textbf{WCS-LT Dataset} is provided by the Wildlife Conservation Society (WCS), and Beery \etal \cite{beery2020iwildcam} split the data by camera location, focusing on predicting unseen camera-trap images.
We use the annotated train split of \cite{beery2020iwildcam}, which contains 217,959 images from 22,111 sequences where only 8,563 sequences include animal species. Here, we use sequences with at least three frames and then select the first three frames according to our observation.  Furthermore, we filter out dominant domain samples to fit the number of recessive domain samples to create a domain-balanced test dataset.

We use 60\% of filtered sequences as the training set and 40\% as the test set, and select only categories with at least one data in each domain and each split (i.e., \emph{train} and \emph{test}).
The training and test set contains 7,416 and 3,990 images, respectively, collected from 211 locations and 34 species represented in the dataset.

\noindent
\textbf{DMZ-LT Dataset} is collected from the Korean Demilitarized Zone (DMZ), which is currently inaccessible due to the ceasefire. The 4,772 sequences consisting of three consecutive frames contain 10 species captured in 99 locations. The two species (i.e., \emph{elk} and \emph{wild boar}) account for 70\% of the entire dataset, resulting in the highly imbalanced label distribution that makes the task challenging. We also filter out dominant domain samples to create a domain-balanced test dataset. Then, we split half of the entire sequences into the training set and the other half into the test set. The training set contains 7,146 images, and the test set contains 5,148 images.

\section{Experiments}

\subsection{Settings}

\noindent
\textbf{Implementation Details.} During training, we set the base learning rate $\eta_{full}$ of the SGD optimizer to 0.01 for WCS-LT and 0.001 for DMZ-LT for 100 epochs, and batch size is set to 48. The $\psi_{full}$ uses $\eta_{full}$, while the learning rate $\eta_{i}$ of each sub-domain expert $\psi_{i}$ follows the Linear Scaling Rule \cite{goyal2017accurate} as

\begin{equation}
    \eta_{i} = \eta_{full} \cdot \frac{\sum_{c \in \mathbb{C}}{n^{c}_{i}}}{\sum_{z \in \mathbb{D}}{\sum_{c \in \mathbb{C}}{n^{c}_{z}}}},
\end{equation}
where $n^{c}_{z}$ is the number of samples for domain $z$ and label $c$.
Input images are resized to 256 $\times$ 256, flipped horizontally with a probability of $\frac{1}{2}$. Moreover, $L^{full}$ updates the backbone and parameters of $\psi_{full}$, and each $L^{i}$ only updates $\psi_{i}$ to alleviate the learning conflict.

\noindent
\textbf{Evaluation Metrics.} We first evaluate the accuracy on many-shot (more than 100 samples), medium-shot (20 $\sim$ 100 samples), and few-shot (less than 20 samples) splits, which are generally evaluated for long-tailed recognition tasks. To better understand the performance of different methods for multiple long-tailed distributions, we calculate the accuracy for major samples (MJs) and minor samples (MNs) separately. We further split $\mathbb{C}$ into a balanced class set and an imbalanced class set according to the ratio of the number of samples in the domain, i.e., the imbalanced class set is defined as
\begin{equation}
\mathbb{C}_{imbal} = \left\{ c \in \mathbb{C}\ |\ \frac{\max_{z \in \mathbb{D}} n^{c}_{z}}{\min_{z \in \mathbb{D}} n^{c}_{z}} \geq 3 \right\}.
\end{equation}
Then, we define the remaining class set as a balanced class set $\mathbb{C}_{bal}$. This results in 32.3\% categories of WCS-LT and 30\% categories of DMZ-LT being in $\mathbb{C}_{imbal}$.
Taken together, we evaluate the accuracy for MJs and MNs of $\mathbb{C}_{bal}$ and $\mathbb{C}_{imbal}$. Note that the average accuracy of MJs and MNs is equal to the total accuracy since we use the domain-balanced test dataset.  

\subsection{Experimental Results}

In this section, we validate our proposed method with comparison to previous long-tailed recognition algorithms \cite{cui2019class, cao2019learning, cai2021ace, lin2017focal}, and experimental results are represented in Table \ref{table:compare}.

For the WCS-LT dataset, our method outperforms other methods for all evaluation metrics except for the few-shot split.
Although ACE \cite{cai2021ace} achieves the best performance on the few-shot split, total accuracy is much lower than baseline since the network is biased toward MJs.
We achieve remarkable improvement on MNs, exceeding the baseline by a margin of 3.3\%p for $\mathbb{C}_{bal}$ and 4.9\%p for $\mathbb{C}_{imbal}$. Interestingly, our method improves even for MJs on $\mathbb{C}_{imbal}$ by 9\%p, which means that domain experts complement each other to potentiate the classification confidence for MJs.

For the DMZ-LT dataset, the difference in accuracy between MJs and MNs of $\mathbb{C}_{bal}$ is about 1\%p, even MNs are more accurate, whereas $\mathbb{C}_{imbal}$ has a difference of more than 50\%p.
These biases towards the dominant domain attenuate the recessive domain prediction, leading to the shrunken decision boundary for the recessive domain. 
Our method exceeds the baseline on MNs for $\mathbb{C}_{imbal}$ by a margin of 28\%p, meaning that the proposed framework resolves the bias even with severe data imbalances.

\subsection{Ablation Study}

We also conduct an ablation study to confirm that each part of our unified framework significantly improves the performance of camera-trap recognition.

\begin{figure}[htb!]
\centering
\resizebox{0.48\textwidth}{!}{

\begin{tabular}{@{}ccccc@{}}
    \includegraphics[width=.4\linewidth, height=.4\linewidth]{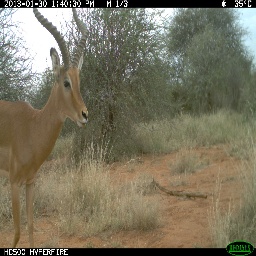} & 
    \includegraphics[width=.4\linewidth, height=.4\linewidth]{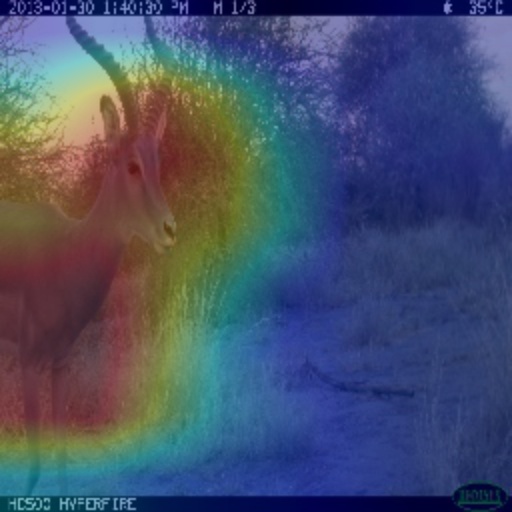} & 
    \includegraphics[width=.4\linewidth, height=.4\linewidth]{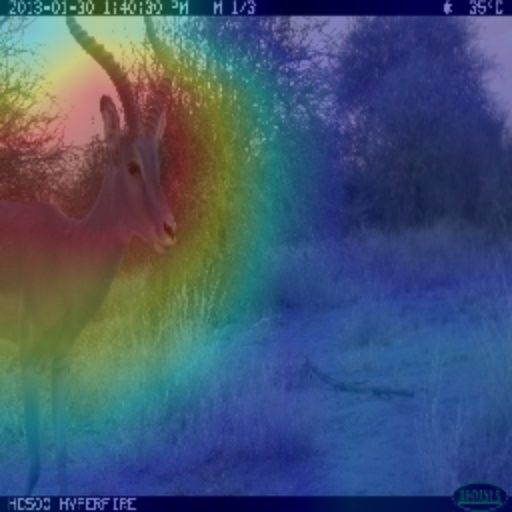} & 
    \includegraphics[width=.4\linewidth, height=.4\linewidth]{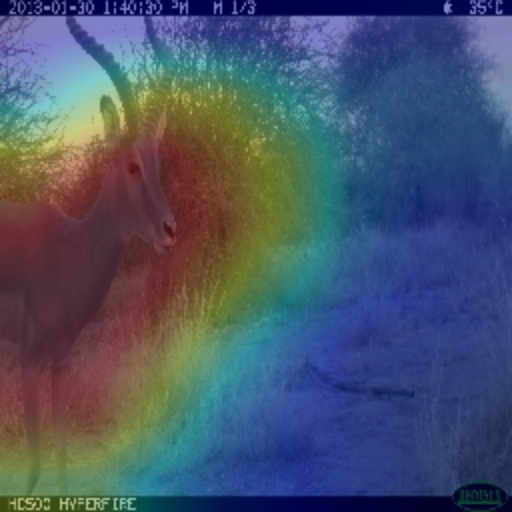} & 
    \includegraphics[width=.4\linewidth, height=.4\linewidth]{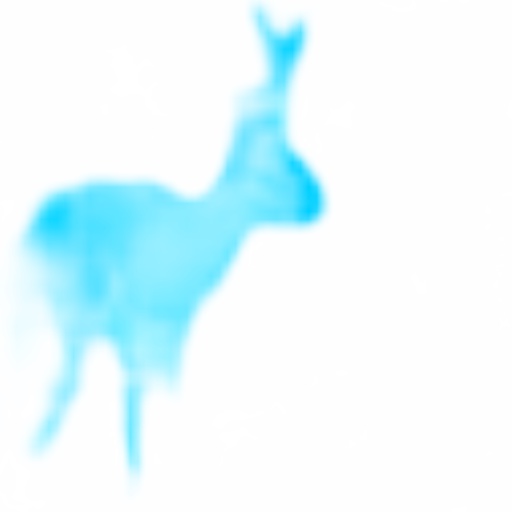} \\
    \includegraphics[width=.4\linewidth, height=.4\linewidth]{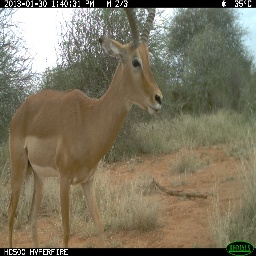} & 
    \includegraphics[width=.4\linewidth, height=.4\linewidth]{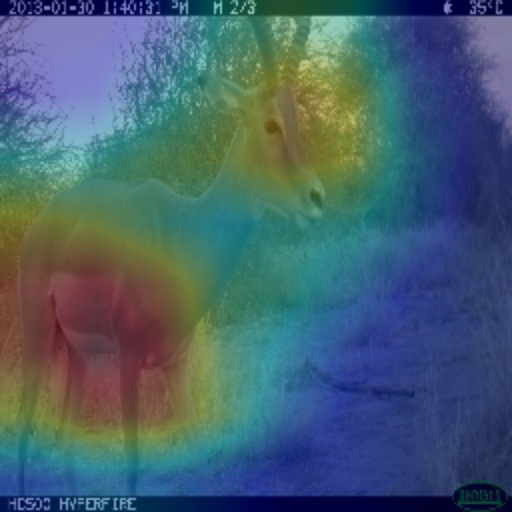} & 
    \includegraphics[width=.4\linewidth, height=.4\linewidth]{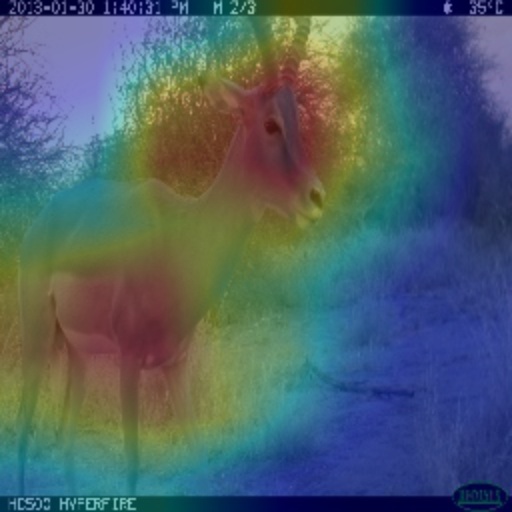} & 
    \includegraphics[width=.4\linewidth, height=.4\linewidth]{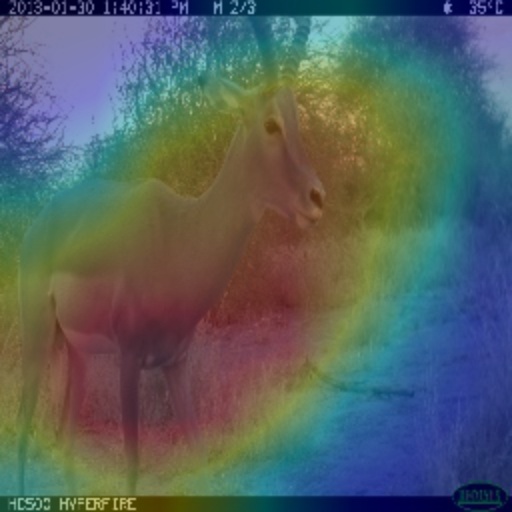} & 
    \includegraphics[width=.4\linewidth, height=.4\linewidth]{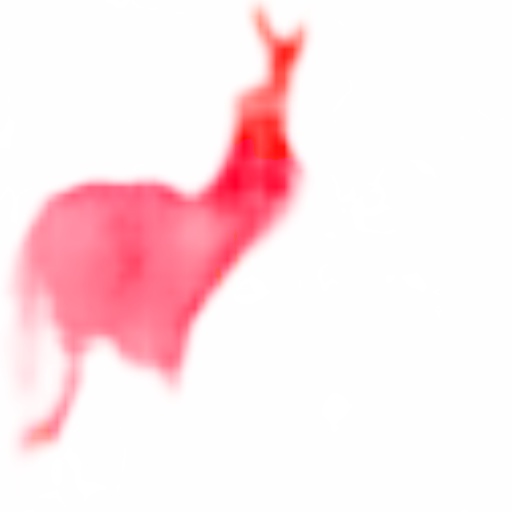} \\
    \includegraphics[width=.4\linewidth, height=.4\linewidth]{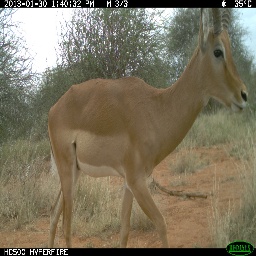} & 
    \includegraphics[width=.4\linewidth, height=.4\linewidth]{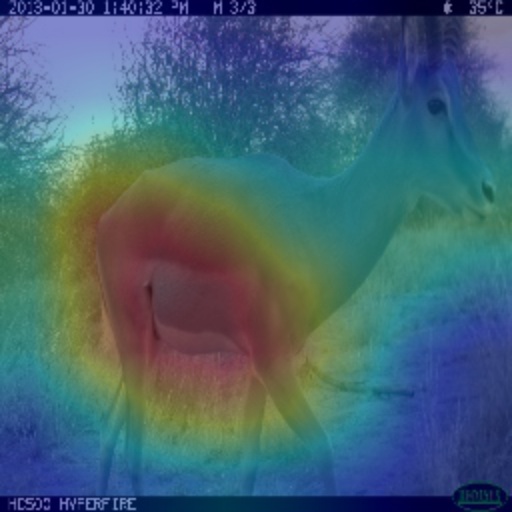} & 
    \includegraphics[width=.4\linewidth, height=.4\linewidth]{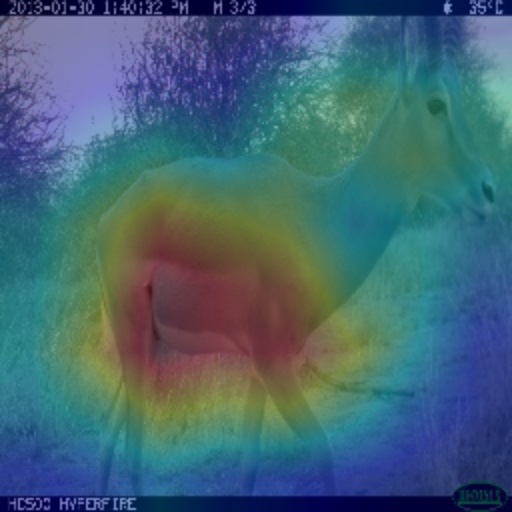} &
    \includegraphics[width=.4\linewidth, height=.4\linewidth]{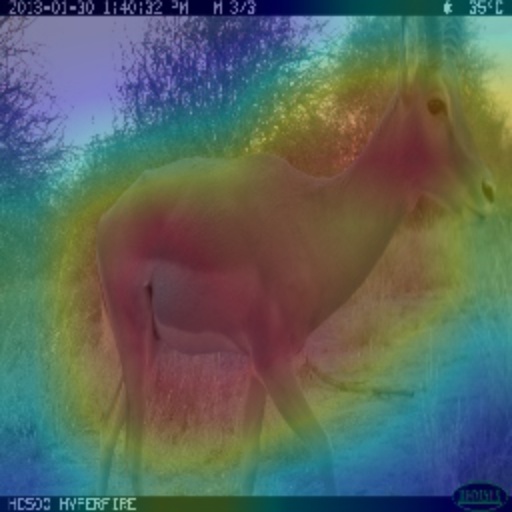} & \\
    \includegraphics[width=.4\linewidth, height=.4\linewidth]{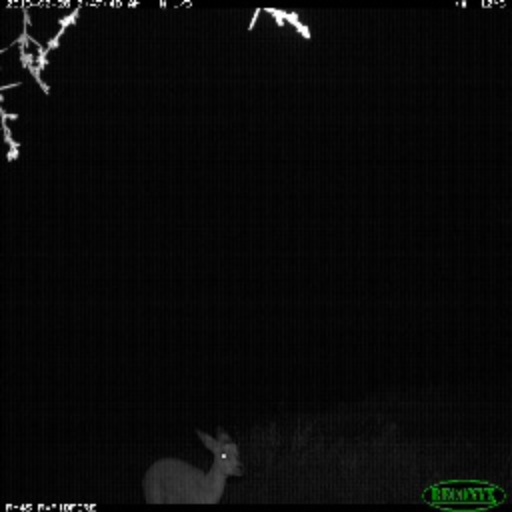} &
    \includegraphics[width=.4\linewidth, height=.4\linewidth]{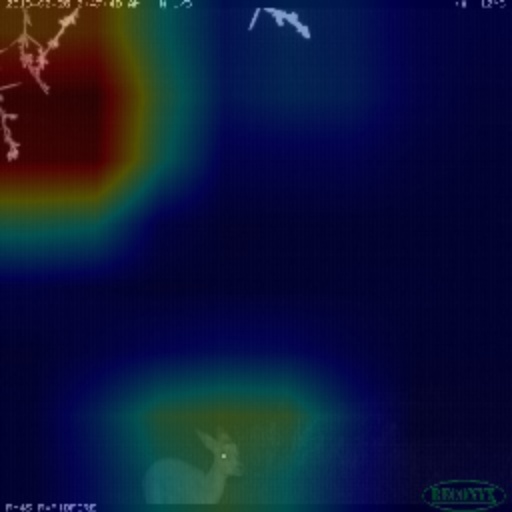} & 
    \includegraphics[width=.4\linewidth, height=.4\linewidth]{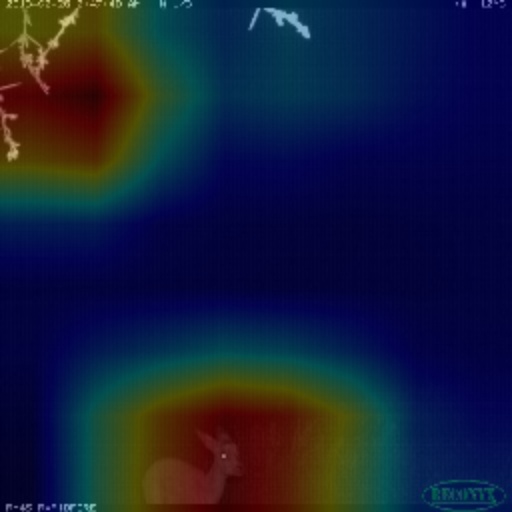} &
    \includegraphics[width=.4\linewidth, height=.4\linewidth]{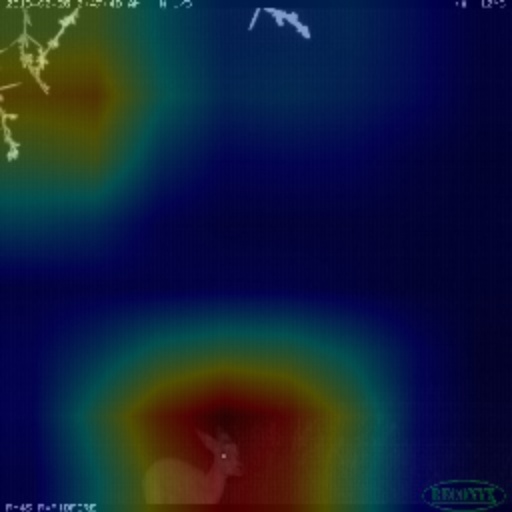} & 
    \includegraphics[width=.4\linewidth, height=.4\linewidth]{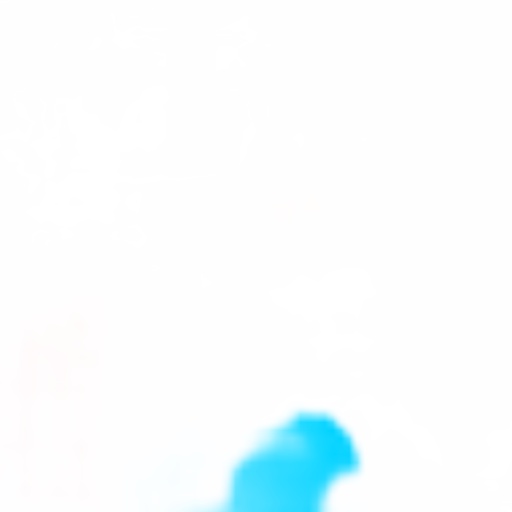} \\  
    \includegraphics[width=.4\linewidth, height=.4\linewidth]{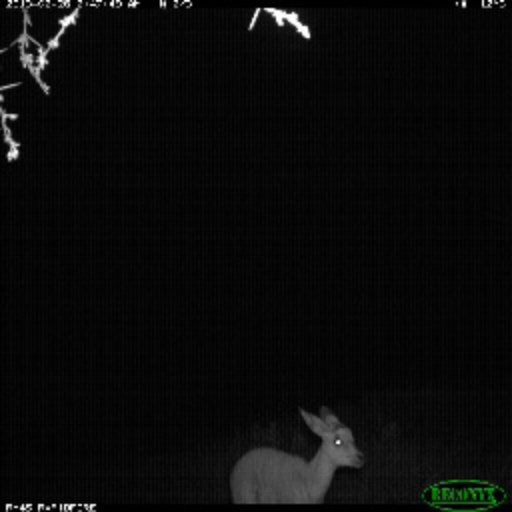} &
    \includegraphics[width=.4\linewidth, height=.4\linewidth]{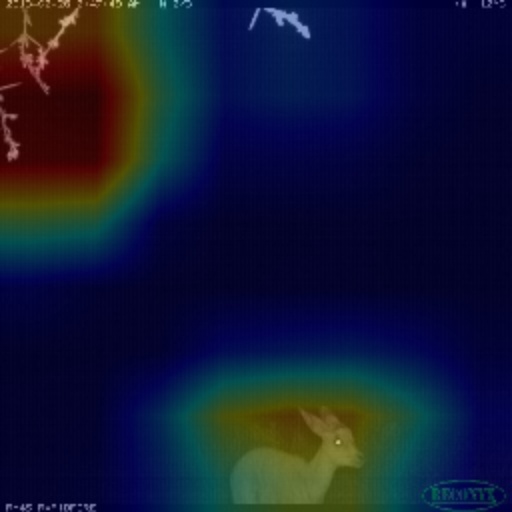} & 
    \includegraphics[width=.4\linewidth, height=.4\linewidth]{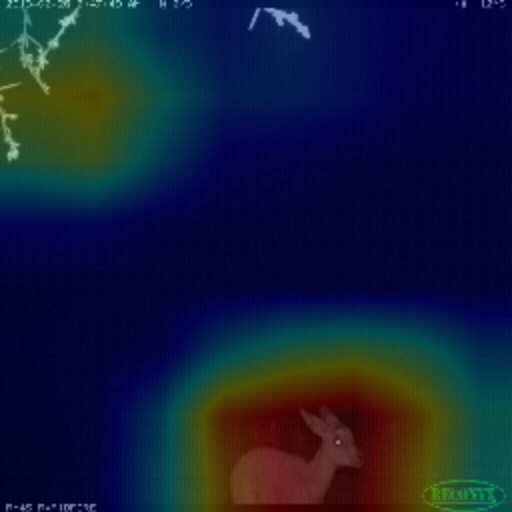} &
    \includegraphics[width=.4\linewidth, height=.4\linewidth]{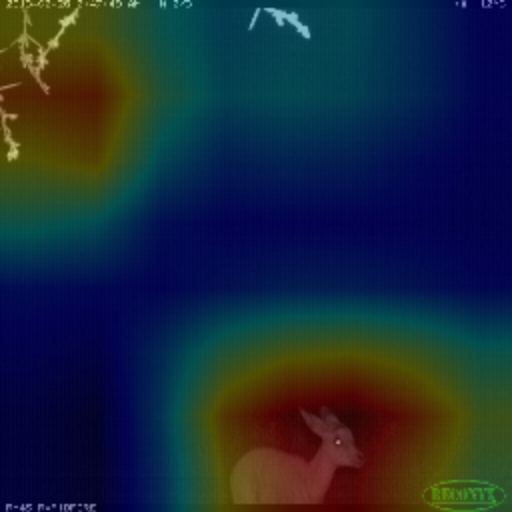} & 
    \includegraphics[width=.4\linewidth, height=.4\linewidth]{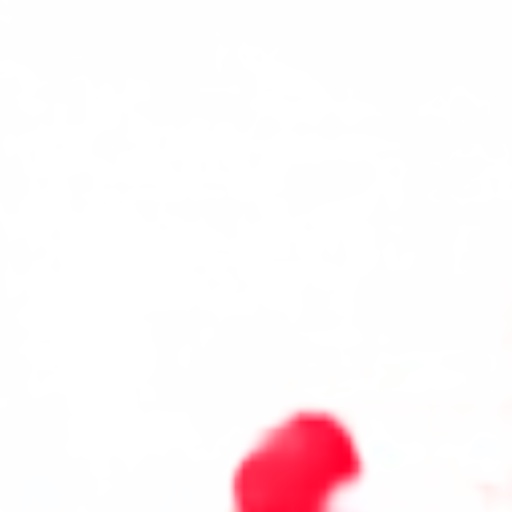} \\
    \includegraphics[width=.4\linewidth, height=.4\linewidth]{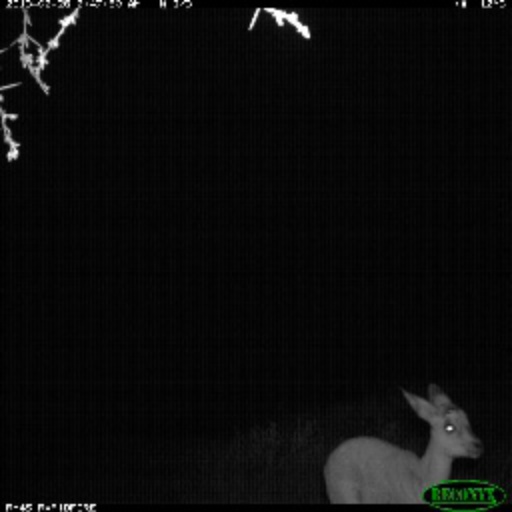} &
    \includegraphics[width=.4\linewidth, height=.4\linewidth]{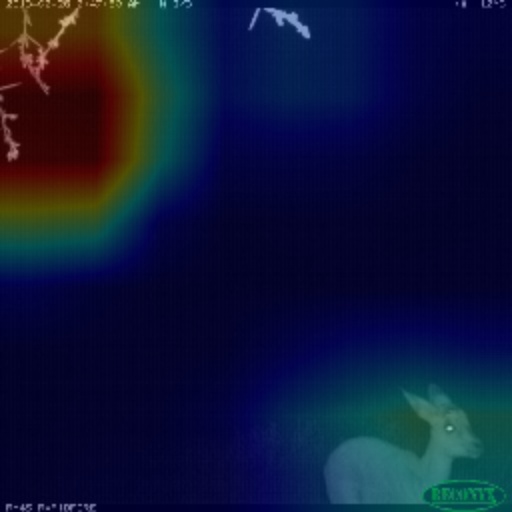} & 
    \includegraphics[width=.4\linewidth, height=.4\linewidth]{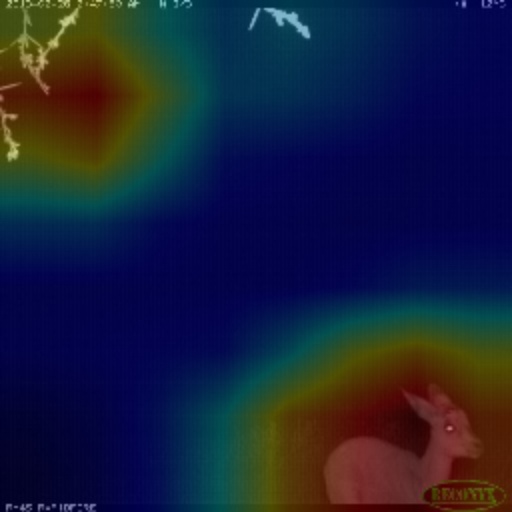} &
    \includegraphics[width=.4\linewidth, height=.4\linewidth]{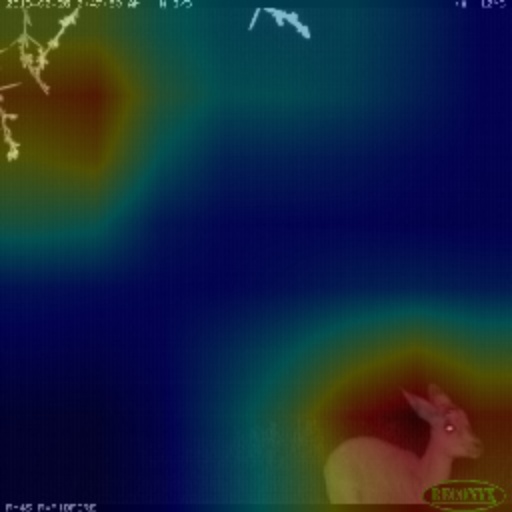} & \\ \\
    {\parbox{.4\linewidth}{\centering {\Large (a) Input Images} }} &
    {\parbox{.4\linewidth}{\centering {\Large (b) Focal \cite{lin2017focal}} }} &
    {\parbox{.4\linewidth}{\centering {\Large (c) Focal w/ FC}}} &
    {\parbox{.4\linewidth}{\centering {\Large (d) Ours (full)} }} & 
    {\parbox{.4\linewidth}{\centering {\Large (e) Flow Maps} }}
  \end{tabular}
 }
\caption{Visualizations of class activation maps and optical flow maps for \emph{"cephalophus silvicultor"}. The upper three rows are for RGB sequences, and the lower three rows are for IR sequences. Also, two consecutive flow maps for each domain represent $f_{2 \to 1}$ and $f_{2 \to 3}$, respectively.}
\vspace{-0.4cm}
\label{fig:qual}
\end{figure}

\begin{table}[htb!]
\centering
\resizebox{1.0\linewidth}{!}{
\begin{tabular}{|c|cc|ccc|}
\Xhline{3\arrayrulewidth}
\multirow{2}{*}{Baseline}   & \multirow{2}{*}{\parbox{1cm}{\centering DE}}  & \multirow{2}{*}{\parbox{1cm}{\centering FC}}  &  \multicolumn{3}{c|}{Top-1 Accuracy (\%)}  \\ \cline{4-6}
                            &               &        &   Major & Minor & All \\ \hline \hline

\multirow{4}{*}{Focal Loss \cite{lin2017focal}}                           &             &               & 80.6 & 79.6 & 80.1 \\ 
                            & \checkmark  &               & 81.2 & 80.2 & 80.7  \\ 
                            &             & \checkmark    & 79.4 & 79.5 & 79.4 \\ 
                            &  \checkmark & \checkmark    & \textbf{81.4} & \textbf{82.4} & \textbf{81.9} \\\Xhline{3\arrayrulewidth}
\end{tabular}%
}
\caption{Ablation study for our proposed method on the WCS-LT dataset. We use a simple long-tailed recognition method \cite{lin2017focal} as the baseline. \textbf{DE} means the domain-experts and \textbf{FC} means the flow consistency loss. Major and Minor metrics are the total accuracy for entire major and minor samples, respectively. Best results in each metric are in \textbf{bold}.}
\label{table2}
\vspace{-0.2cm}
\end{table}

Table \ref{table2} shows the quantitative result to verify the effect of domain experts and the flow consistency loss.
Domain experts improve the classification performance for all evaluation metrics, and further improvements are achieved when the flow consistency loss is applied to experts.
Surprisingly, the flow consistency loss does not address the data imbalances in baseline, but synergizes with domain experts to considerably improve the performance.
Collectively, our unified framework improves 1.8\%p for MJs and 2.8\%p for MNs compared to the focal loss \cite{lin2017focal}.

Qualitative results in Fig. \ref{fig:qual} indicate the flow consistency loss regulates the classifier to pay more attention to moving objects. Specifically, our method focuses on the general contextual information of the animal in RGB images, and the class activation map of our method mainly highlights the moving animal while the baseline focuses on the brightest object in a situation with low light conditions. In this regard, qualitative and quantitative results strengthen the position that complementary domain experts have better discriminability and are superior to the baseline.

\section{Conclusion}

In this work, we have proposed a unified framework for long-tailed camera-trap recognition and introduced two benchmark datasets, WCS-LT and DMZ-LT. The main contribution is that domain experts are balanced through the loss re-weighting and complement each other to provide the domain-balanced decision boundaries. We also design the flow consistency loss that experts pay more attention to moving objects in camera-trap images. We believe that our datasets will contribute to camera-trap studies. In the future, we plan to extend our framework for domain generalization tasks, considering long-tailed distributions for diverse domains.

\section{Acknowledgement}

This work was supported by the National Research Foundation of Korea (NRF) grand founded by the Korea Government (MSIT) (NRF-2018R1A5A7025409)

\bibliographystyle{IEEEbib}
\bibliography{icme2022template}

\end{document}